\documentclass{article}

\usepackage{microtype}
\usepackage{graphicx}
\usepackage{subfigure}
\usepackage{booktabs} 
\usepackage{amsmath}
\usepackage{amsfonts}
\usepackage{amssymb}
\usepackage{multirow}

\usepackage{hyperref}

\usepackage[accepted]{icml2021}

\icmltitlerunning{Extreme Precipitation Seasonal Forecast Using a Transformer Neural Network}

\begin{document}

\twocolumn[
\icmltitle{Extreme Precipitation Seasonal Forecast Using a Transformer Neural Network}



\icmlsetsymbol{equal}{*}

\begin{icmlauthorlist}
\icmlauthor{Daniel Salles Civitarese}{equal,ibm}
\icmlauthor{Daniela Szwarcman}{equal,ibm}
\icmlauthor{Bianca Zadrozny}{ibm}
\icmlauthor{Campbell Watson}{ibm}
\end{icmlauthorlist}

\icmlaffiliation{ibm}{IBM Research, Rio de Janeiro, Brazil}

\icmlcorrespondingauthor{Daniel Salles Civitarese}{sallesd@br.ibm.com}

\icmlkeywords{
    transformer network,
    extreme events,
    precipitation,
    seasonal forecast,
    quantile regression
}

\vskip 0.3in
]



\printAffiliationsAndNotice{\icmlEqualContribution} 

\begin{abstract}
An impact of climate change is the increase in frequency and intensity of extreme precipitation events. However, confidently predicting the likelihood of extreme precipitation at seasonal scales remains an outstanding challenge. Here, we present an approach to forecasting the quantiles of the maximum daily precipitation in each week up to six months ahead using the temporal fusion transformer (TFT) model. Through experiments in two regions, we compare TFT predictions with those of two baselines: climatology and a calibrated ECMWF SEAS5 ensemble forecast (S5). Our results show that, in terms of quantile risk at six month lead time, the TFT predictions significantly outperform those from S5 and show an overall small improvement compared to climatology. The TFT also responds positively to departures from normal that climatology cannot.

\end{abstract}

\section{Introduction}
\label{intro}

Extreme weather and climate events can severely impact communities, infrastructure, agriculture and other natural and human systems.
The observed increase in frequency and intensity of extreme events has been directly linked to anthropogenic climate change, and the trend is expected to continue in the coming years \cite{Extremes2019}.
Improving our understanding of the impacts of climate change on weather and climate extremes and developing early warning systems for better preparedness is of paramount importance.

This short paper is concerned with extreme precipitation prediction, which can cause flooding, crop damage, and widespread disruption to ecosystems.
The ability to confidently predict the likelihood of extreme precipitation at sub-seasonal to seasonal scales remains a significant challenge \cite{King2020}. Recent works have shown that statistical machine learning (ML) models can outperform state-of-the-art dynamical models when predicting extreme events several months ahead (e.g., \citet{Chantry2021}, \citet{Cohen2019}). These ML models tend to rely on slowly-changing variables, such as soil moisture and El Ni{\~{n}}o-Southern Oscillation (ENSO) indices \cite{fernando2019process}.
Although most of these variables are publicly available, the degree of influence of each one regarding precipitation prediction varies in space and time (e.g., \citet{Strazzo2019}).
Properly accounting for these effects in a ML model is a complex task.

In this work, we present an ML approach to forecast the \textit{maximum daily precipitation in each week} up to six months ahead. Our approach uses the temporal fusion transformer (TFT) model, which combines multi-horizon forecasting with specialized components to select relevant inputs and suppress unnecessary features \cite{lim2019temporal}.
Of particular interest to both researchers and end-users, the TFT produces multiple quantiles.
These quantiles offer a global view of the interrelations between the input and output \cite{davino2013quantile} and can help risk management by indicating the likely best and worst-case values of the target variable \cite{lim2019temporal}.
To the Authors' knowledge, this is the first time the TFT has been used to forecast extreme weather.

\section{Methodology}
\label{method}

We define our task as a quantile forecast for precipitation in multiple locations.
The structure is similar to the one presented in \cite{lim2019temporal}, where the authors consider three kinds of input: (a) static $\mathbf{s} \in \mathbb{R}^{m_s}$, (b) known future $\mathbf{x} \in \mathbb{R}^{m_x}$, and (c) historical $\mathbf{z} \in \mathbb{R}^{m_z}$ (Figure~\ref{fig:tft} -- yellow, pink, and blue, respectively).
Static information does not change through time, e.g., location or altitude.
Examples of known future information are the month-of-the-year and external predictions for the target and other variables, such as temperature.
Although predictions are not ``known information'' and may have errors, their use follows the same procedure as future inputs.
Finally, historical inputs are the actual past values for the target series or any other exogenous one, e.g., temperature or soil moisture.

We produce weekly forecasts for each location in the form $\hat{y}_{lat, lon}(q, t, \tau)$, where each time-step $t \in [0, T]$ refers to a specific week in the time series, and $\tau \in [1, 26]$ is the lead time in weeks.
The index $(lat, lon) = i$ refers to the location in the globe, and it is associated with a set of static inputs $\mathbf{s}_{i}$, as well as the time-dependent inputs $\mathbf{z}_{i, t}$ and $\mathbf{x}_{i, t}$.

We predict the quantiles 0.1, 0.5, and 0.9 of the maximum daily rainfall for each week up to six months ahead in our setup.
Each quantile is represented by $\hat{y}_{i}(q, t, \tau) = f_{q}(\tau, y_{i, t-k:t}, z_{i, t-k:t}, x_{i, t-k:t+\tau}, s_{i})$
Although all quantiles refer to the maximum daily rainfall in a week, we pay closer attention to quantile 0.9 to focus on extreme events.

\subsection{TFT model}

The primary motivation of using the TFT model for our task is its ability to handle multiple types of variables, i.e., static, historical, and future.
The model also provides other features, such as simultaneous outputs for all $\tau$ time steps and a modified multi-head attention mechanism that facilitates the interpretation of the outputs.
We intend to use the latter in future work.

\begin{figure}[h]
\vskip 0.1in
\begin{center}
\centerline{\includegraphics[width=\columnwidth]{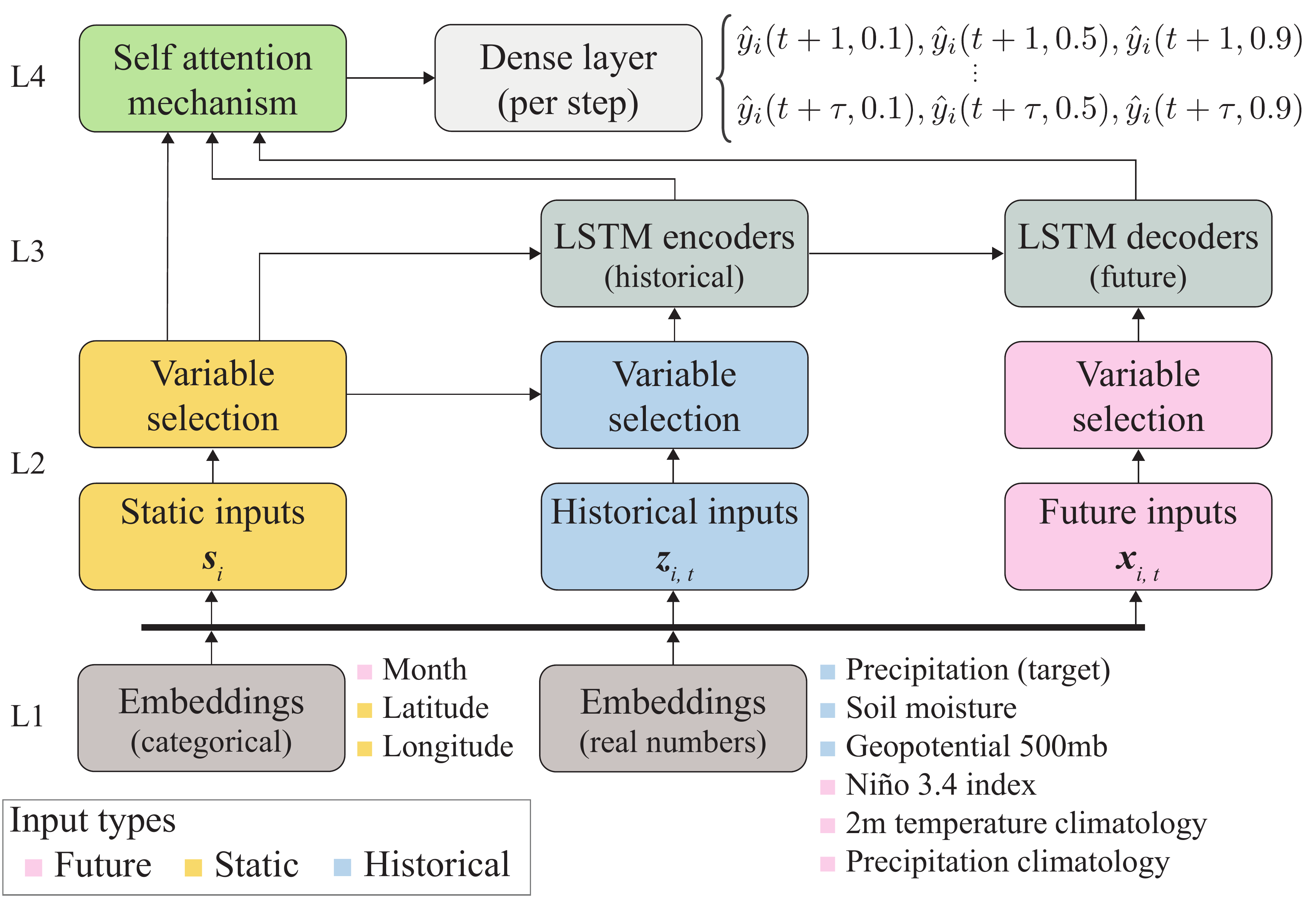}}
\caption{Overview of the TFT architecture. It shows the types of inputs: static (yellow), historical (blue), and future (pink). Main components are: variable selection, LSTM encoders and decoders, and the self-attention mechanism.}
\label{fig:tft}
\end{center}
\vskip -0.2in
\end{figure}

Figure~\ref{fig:tft} shows a summarized version of the TFT structure.
Level L1 creates input embeddings from both categorical and continuous values.
In Section~\ref{exps}, we describe the input variables used in our experiments.
Level L2 processes each input type individually and selects relevant input variables at each step.
Static encoders (yellow) integrate time-invariant features into the network to condition temporal dynamics.
The sequence-to-sequence LSTM module learns short-term temporal relationships, whereas the multi-head attention block on level L4 captures long-term dependencies.

As in \cite{lim2019temporal}, we train the TFT model by jointly minimizing the quantile loss, summed across all quantile outputs:
\begin{equation}
    \mathcal{L}(\Omega, \textbf{W}) = \sum_{y_t \in \Omega}{\sum_{q \in \mathcal{Q}}{\sum_{\tau = 1}^{\tau_{max}}{\frac{QL \left(y_t, \hat{y}(q, t-\tau, \tau), q \right)}{M \tau_{max}}}}}
    \label{eq:24}
\end{equation}
\begin{equation}
    QL(y, \hat{y}, q) = q(y-\hat{y})_{+} + (1-q)(\hat{y}-y)_{+}
    \label{eq:25}
\end{equation}
\noindent where $\Omega$ is the domain of the training data, which contains $M$ samples, $\textbf{W}$ refers to the TFT weights, $\mathcal{Q}$ is the set of output quantiles, and $(.)_+ = max(0, .)$.

The normalized quantile loss (q-risk) is used for validation and testing:
\begin{equation}
    \text{q-risk} = \frac{2\sum_{y_t \in \tilde{\Omega}}{\sum_{\tau=1}^{\tau_{max}}{QL\left( y_t, \hat{y}(q, t - \tau, \tau), q \right)}}}{\sum_{y_t \in \tilde{\Omega}}{\sum_{\tau=1}^{\tau_{max}}}\left| t_t \right|}
    \label{eq:26}
\end{equation}
\noindent where $\tilde{\Omega}$ is the domain of the validation or test samples.

\subsection{Baseline models}

\textbf{Climatology.} Climatology is a simple model that is commonly used as a baseline for seasonal forecast skill. Here, it is computed from historical values (1981-2010) of the variable of interest and depends on the selected temporal aggregation, e.g., monthly mean. We follow the steps in Algorithm \ref{alg:climo} to compute the climatology quantiles for precipitation ($agg = max$, $op = quantiles$), creating predictions for each week of the year (1 to 53) for any given year.

\begin{algorithm}[ht]
   \caption{Climatology - week}
   \label{alg:climo}
\begin{algorithmic}
   \STATE {\bfseries Input:} data (1981-2010) $x$, aggreg. $agg$, operation $op$
   \STATE {\hspace{30pt}$agg \in [max, mean, min, ...]$}
   \STATE {\hspace{30pt}$op \in [mean, quantiles, max, ...]$}
   \STATE $week\_x \gets$ Aggregate $x$ weekly, using the $agg$ function
   \STATE Initialize array $climo$, size = 53
   \FOR{$i$ = 1 {\bfseries to} 53}
   \STATE $week\_values \gets$ values for all weeks $i$ in $week\_x$
   \STATE $climo[i] \gets op(week\_values)$
   \ENDFOR
\end{algorithmic}
\end{algorithm}

\textbf{Calibrated ECMWF SEAS5.} We also compare our predictions to those from a calibrated ECMWF SEAS5 (hereafter S5) seasonal forecast 50-member ensemble \cite{Crawford2019}. To compute the forecasted quantiles from the S5 ensemble, we first aggregate the data to obtain the weekly maximum precipitation and then calculate the quantiles across the 50 members for each week of interest.

\section{Experiments}
\label{exps}

\subsection{Data}

We selected two regions to investigate: Rio de Janeiro (Brazil) and Florida (USA). Besides precipitation, we consider as inputs to the network: 2 m temperature, soil moisture\footnote{volumetric soil water from 0 to 7 cm of depth}, and geopotential height at 500 mbar (all from the location of the forecasts). This choice was based on past work that used these variables in the context of rainfall prediction \cite{fernando2019process, Xu2020}.

The literature also indicates that precipitation is considerably affected by ENSO in Rio and Florida \cite{Arguez2019, Grim2009}. To account for ENSO, we consider the Niño 3.4 index prediction as a future input. Currently, we are using Niño 3.4 index observations\footnote{\url{http://www.jamstec.go.jp}} as if they were predictions to train the networks, and we plan to replace these with predictions from the S5 ensemble.

The data selected for observed precipitation was CHIRPSv2, \cite{chirps}, a daily gridded dataset with a spatial resolution of 0.05$^{\circ}$. For the other climate variables, we used ERA5 reanalysis data for surface \cite{era5surface} and pressure \cite{era5pressure} variables. ERA5 is also a gridded dataset but with resolution of 0.25$^{\circ}$. We performed a spatial max-pooling operation to convert the CHIRPSv2 data to the resolution of ERA5 (5$\times$5 window), which guarantees that the highest precipitation values for each 5$\times$5 window will not be smoothed out.

Precipitation, geopotential, and soil moisture are provided as historical inputs to the network: 26 weeks of past values. We consider the weekly mean for soil moisture and geopotential. For temperature, we compute the weekly mean climatology and provide it as a prediction. The precipitation climatology ($agg = max$, $op = mean$) is also given as a future input.

Finally, we added some categorical variables: month as a future input, and latitude and longitude as static inputs. The idea behind these static inputs is to provide some spatial information to the model. Figure \ref{fig:tft} lists all the inputs.

\subsection{Training}

\begin{figure}[ht]
\vskip 0.1in
\begin{center}
\centerline{\includegraphics[width=\columnwidth]{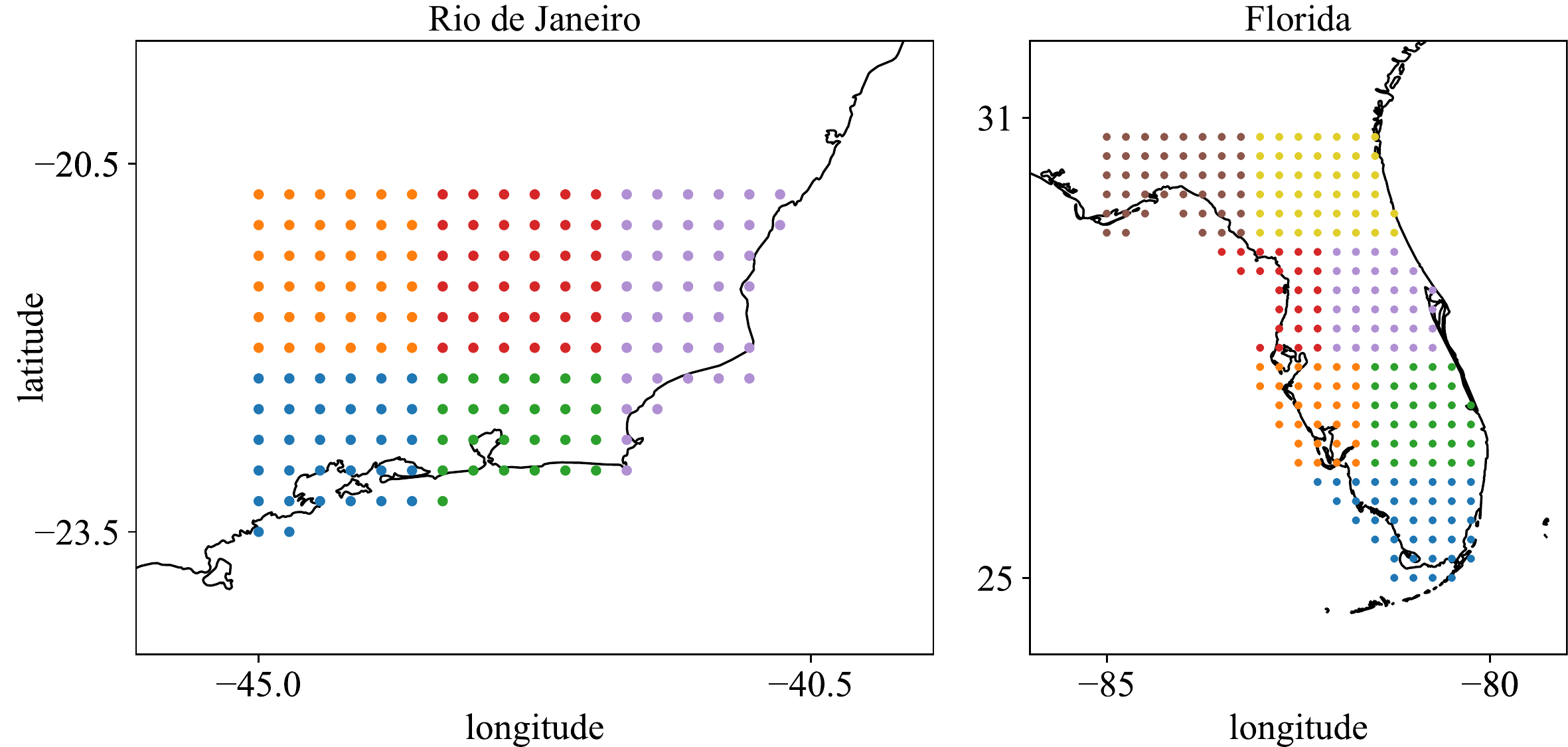}}
\caption{Selected regions and grid points in Rio de Janeiro (left) and Florida (right). The grid points are divided into smaller sub-regions, shown in different colors.}
\label{fig:groups}
\end{center}
\vskip -0.2in
\end{figure}

As the regions of interest contain numerous points (170 in Rio and 248 in Florida), we divide each region into smaller sub-regions and train a different TFT model for each. The sub-regions have around 20-45 grid points each, a decision based on preliminary results. 
Figure \ref{fig:groups} shows the selected grid points in Rio (3$^{\circ}\times$ 4.5$^{\circ}$) and Florida (6$^{\circ}\times$ 5$^{\circ}$).

We split the data into three non-overlapping ranges to define our training (1981-2010), validation (2011-2014), and test (2015-2019) sets. We perform hyperparameter optimization with random search following the same scheme presented by the TFT authors. The optimization is carried out for 60 iterations, and the list of hyperparameters includes \emph{dropout rate}, \emph{hidden layer size}, \emph{minibatch size}, \emph{learning rate} (Adam optimizer), \emph{maximum gradient norm}, and \emph{number of heads}.
Each TFT is trained with one Nvidia V100 GPU for 100 epochs (early-stopping patience of 5 epochs). The optimization takes 3 to 4 hours to run with this configuration.

\subsection{Evaluation}

The TFT makes predictions for weeks 1 to 26, and here we present results only for $\tau$ = 26, which means that we remove the summations over $\tau$ in Eq.~\ref{eq:26} and replace $\tau$ by $\tau_{26}$. The lead time of 26 weeks is both challenging and directly relevant to a range of stakeholder applications. We compare the TFT predictions with both climatology and S5 predictions. Since the S5 spatial resolution is 0.4$^{\circ}$, we use nearest neighbors interpolation to re-grid the data to the resolution of our predictions (0.25$^{\circ}$).

\section{Results and Discussion}
\label{results}

\begin{figure}[h]
\vskip 0.1in
\begin{center}
\centerline{\includegraphics[width=\columnwidth]{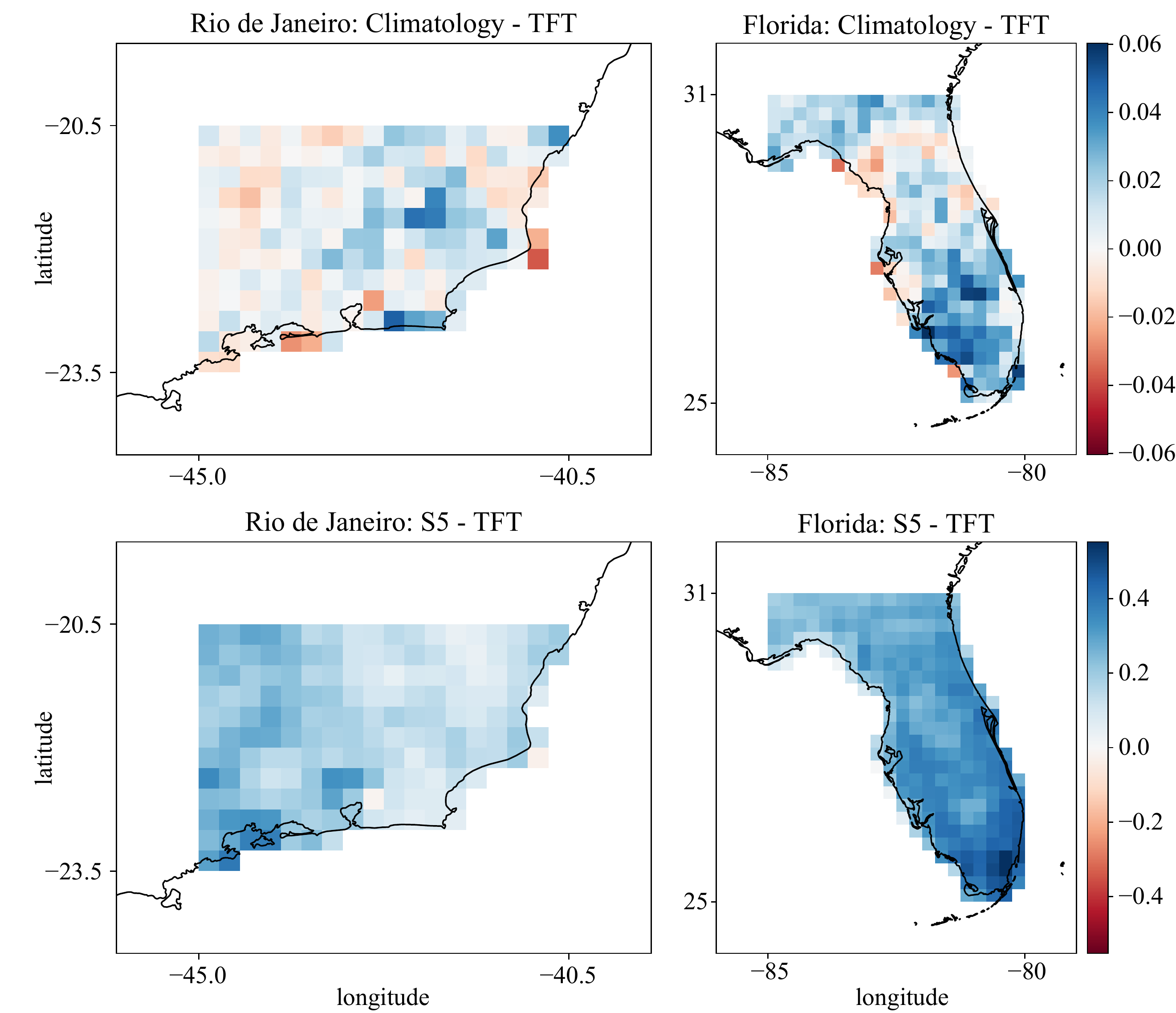}}
\caption{Q-risk comparison in Rio and Florida for quantile 0.9. Top: q-risk(Climo) - q-risk(TFT). Bottom: q-risk(S5) - qrisk(TFT).}
\label{fig:q09_diff}
\end{center}
\vskip -0.2in
\end{figure}

Figure \ref{fig:q09_diff} shows the q-risk difference between the baseline models and TFT for quantile 0.9: climatology - TFT (top) and S5 - TFT (bottom). 
Positive values indicate the TFT shows superior results (lower values of q-risk are better). Compared to climatology, TFT has lower q-risks for most locations. Compared to S5, TFT exhibits considerably lower q-risk for all points – a substantial improvement.

In Table \ref{tab:q-risk_differences}, we present the average q-risk difference between the reference models and TFT. Regarding climatology, TFT shows better results for all quantiles in Rio and quantile 0.9 in Florida. The average q-risk difference between S5 and TFT is significant for all quantiles in both locations, especially for quantile 0.9, which is our main focus.

\begin{table}[t]
\caption{Average q-risk difference for each quantile.}
\label{tab:q-risk_differences}
\vskip 0.15in
\begin{center}
\begin{tabular}{@{}clccc@{}}
\toprule
Comparison & Region & 0.1 & 0.5 & 0.9 \\ \midrule
\multirow{2}{*}{$\frac{\text{climo - TFT}}{\text{climo}}$} & Rio & 2.45\% & 0.90\% & 1.08\% \\
 & Florida & -2.16\% & -0.41\% & 3.70\% \\ \midrule
\multirow{2}{*}{$\frac{\text{S5 - TFT}}{\text{S5}}$} & Rio & 3.71\% & 11.18\% & 29.54\% \\
 & Florida & 5.70\% & 16.15\% & 41.87\% \\ \bottomrule
\end{tabular}
\end{center}
\vskip -0.1in
\end{table}

Although the average q-risks for quantile 0.9 seems to show a small improvement compared to climatology, we highlight that, for seasonal time scale, climatology is exceptionally hard to improve upon \cite{Vitart2018}.
Furthermore, since climatology generates fixed predictions for all years, it cannot capture sudden changes or departures from normal.

\begin{figure}[h]
\vskip 0.1in
\begin{center}
\centerline{\includegraphics[width=0.93\columnwidth]{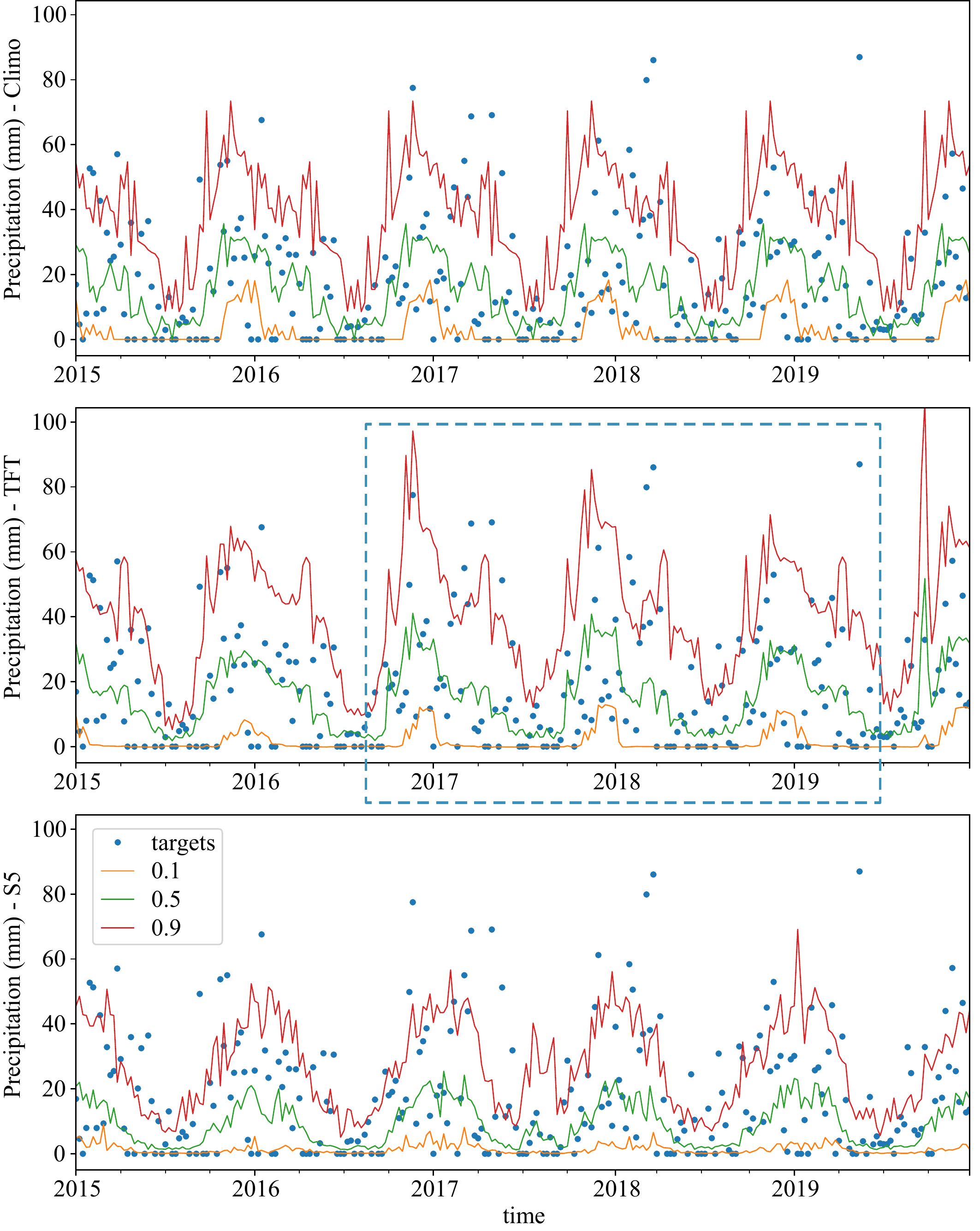}}
\caption{Predictions and targets for the test set in the location (lat -21.5, lon -41.75) in Rio. Top: Climatology. Middle: TFT. Bottom: S5. In this location, the q-risk difference for quantile 0.9 between climo and TFT was 1.9\%, and S5 and TFT was 15.8\%.}
\label{fig:predictions}
\end{center}
\vskip -0.2in
\end{figure}

Figure \ref{fig:predictions} illustrates the quantile predictions and target values of the test set in a single location in Rio de Janeiro for all models. With a relatively small q-risk improvement, one could argue that TFT is mimicking climatology. However, the TFT predictions change considerably throughout the years, contrasting with the fixed climatology behavior.
We draw attention to the TFT predictions for quantile 0.9 inside the dashed box: TFT raised the quantile level during weeks of very heavy rainfall while also responding to the periods of decreased precipitation. 

\section{Conclusions}
\label{conclusions}

In this work, we used the TFT network to predict the maximum daily precipitation quantiles in a week out to a lead time of 6 months.
The TFT generated significantly improved q-risks compared to the S5 model and marginal improvements compared to climatology.
Comparing the 0.9 quantile prediction in one location in Rio, we have shown that TFT can accurately raise the quantile level and also respond to changes that climatology cannot.
These results indicate that TFT has an interesting potential for extreme rainfall forecasting on the sub-seasonal to seasonal scale.

In the following steps, we want to modify the model's input to support 2D spatial information. That would allow us to encode neighborhood knowledge into the process.
Additional pre-processing, such as POD or graph networks, could also capture teleconnections and include sparse datasets, such as sea surface temperature that is only available over the ocean.
We also want to use the interpretable multi-head attention block to identify connections between the input variables and extreme rainfall.

The TFT has the ability to handle both historical data and predicted future data, providing a ready-to-use architecture for performing hybrid statistical-dynamical predictions.
We intend to incorporate other input variables, such as dynamical model predictions, to investigate if they can improve TFT results.

\bibliography{main}
\bibliographystyle{icml2021}

\end{document}